\PassOptionsToPackage{table}{xcolor}
\documentclass[numbers]{article} %
\usepackage{tccml_iclr2025_conference,times}

\usepackage{amsmath,amsfonts,bm}

\def\eqref#1{equation~\ref{#1}}

\def\1{\bm{1}}

\DeclareMathAlphabet{\mathsfit}{\encodingdefault}{\sfdefault}{m}{sl}
\SetMathAlphabet{\mathsfit}{bold}{\encodingdefault}{\sfdefault}{bx}{n}

\usepackage{xcolor}  
\usepackage{url}
\usepackage[english]{babel}      
\usepackage[T1]{fontenc}         
\usepackage{etoolbox}            
\usepackage{float}                       
\usepackage{graphicx}            
\graphicspath{{Figures/}}        
\usepackage[lined, boxed, linesnumbered, ruled]{algorithm2e} 
\usepackage{lineno,hyperref}     
\usepackage{epstopdf}            
\usepackage{bbding}              
\usepackage{lipsum}            
\usepackage{multirow}           
\usepackage{makecell} 
\usepackage{caption}
\usepackage{picinpar}            
\usepackage{enumerate}                                
\usepackage{subcaption}           
\usepackage{comment}           
\usepackage{hyperref}          
\usepackage{amsmath}             
 
\usepackage{algorithmic}         
\usepackage{balance}             
\usepackage{textcomp}  
\usepackage{pifont}             
\usepackage{multirow}          
\usepackage{color}              
\usepackage{lscape}      
\usepackage{hhline}
\usepackage{longtable,booktabs}  
\usepackage{enumitem}
\usepackage{appendix} 
\usepackage{minitoc}

\definecolor{RowColor}{rgb}{0.97, 0.97, 1}

\newcommand{\cmark}{\textcolor[rgb]{0.004, 0.663, 0}{\ding{51}}} 
\newcommand{\xmark}{\textcolor{red}{\ding{55}}}

\title{Segregation and Context Aggregation Network for Real-time Cloud Segmentation 
}

\renewcommand{\thefootnote}{\fnsymbol{footnote}}
\author{\unskip
  \textbf{Yijie Li}\textsuperscript{1,\footnotemark[1]}, 
  \textbf{Hewei Wang}\textsuperscript{1,\footnotemark[1]}, 
  \textbf{Jiayi Zhang}\textsuperscript{2}, 
  \textbf{Jinjiang You}\textsuperscript{1}, 
  \textbf{Jinfeng Xu}\textsuperscript{3}, 
  \textbf{Puzhen Wu}\textsuperscript{4}, \\
  \textbf{Yunzhong Xiao}\textsuperscript{1}, 
  \textbf{Soumyabrata Dev}\textsuperscript{5, \footnotemark[2]}  
  \\
  \textsuperscript{1}Carnegie Mellon University \quad
  \textsuperscript{2}University of Nottingham\\
  \textsuperscript{3}The University of Hong Kong \quad
  \textsuperscript{4}Cornell University \quad
  \textsuperscript{5}University College Dublin
}

\iclrfinalcopy 
\begin{document}

\maketitle
\footnotetext[1]{Equal contribution.}
\footnotetext[2]{Corresponding authors: Soumyabrata Dev (soumyabrata.dev@ucd.ie)}

\renewcommand{\thefootnote}{\arabic{footnote}}

\begin{abstract}
Cloud segmentation from intensity images is a pivotal task in atmospheric science and computer vision, aiding weather forecasting and climate analysis. Ground-based sky/cloud segmentation extracts clouds from images for further feature analysis. Existing methods struggle to balance segmentation accuracy and computational efficiency, limiting real-world deployment on edge devices, so we introduce SCANet, a novel lightweight cloud segmentation model featuring Segregation and Context Aggregation Module (SCAM), which refines rough segmentation maps into weighted sky and cloud features processed separately. SCANet achieves state-of-the-art performance while drastically reducing computational complexity. SCANet-large (4.29M) achieves comparable accuracy to state-of-the-art methods with 70.9\% fewer parameters. Meanwhile, SCANet-lite (90K) delivers 1390 fps in FP16, surpassing real-time standards. Additionally, we propose an efficient pre-training strategy boosting performance even without ImageNet pre-training. Our code is publicly available at: \url{https://github.com/Att100/SCANet}.
\end{abstract}


\section{Introduction}
Understanding cloud-sky relationships is crucial for climate modeling, solar energy forecasting, and extreme weather prediction. Advances in computer vision and machine learning have improved meteorology estimation~\citep{dev2019estimating, 10845086, 7991380, akrami2022graph} and weather prediction~\citep{manandhar2019data, 10443636, wang2021day, mcnicholl2021evaluating}, offering insights into cloud status. While satellites provide valuable cloud data, they are costly and storage-intensive. Ground-based sky/cloud segmentation~\citep{jain2021extremely, dev2015design, dev2018high, zhang2022novel}, supported by datasets like SWIMSEG~\citep{dev2016color}, SWINSEG~\citep{dev2017nighttime}, and SWINySEG~\citep{dev2019cloudsegnet}, offers a cost-effective, high-resolution alternative, enhancing climate applications.

Sky/cloud segmentation, a binary semantic task, has evolved with fully convolutional networks (FCN)~\citep{long2015fully}. Real-time segmentation strategies include (a) lightweight backbones and decoders, e.g., DeepLab~\citep{chen2014semantic, chen2017deeplab}, and (b) encoder-decoder architectures like ICNet~\citep{zhao2018icnet} and BiseNet~\citep{yu2018bisenet, yu2021bisenet}. However, existing methods either lose information due to attention mechanisms or suffer from slow inference, limiting real-time applications. A detailed discussion of related works, including their limitations, can be found in Appendix~\ref{sec:appendixrelatedwork}, further motivating the design of SCANet. 

To address these challenges, we propose SCANet, a lightweight yet effective model integrating MobileNetV2~\citep{sandler2018mobilenetv2} and EfficientNet-B0~\citep{tan2019efficientnet} with a Segregation and Context Aggregation module (SCAM). SCANet, following a U-Net~\citep{ronneberger2015u} structure, refines cloud-sky features via SCAM decoders. Unlike prior methods, SCAM enhances feature separation and aggregation, while supervision at the last three stages accelerates convergence.

By improving segmentation accuracy and efficiency, SCANet supports climate modeling, renewable energy forecasting, and extreme weather monitoring, demonstrating a direct pathway from machine learning to climate impact.  The main contributions of our SCANet are twofold:

\begin{itemize}
    \item SCANet, a lightweight CNN-based model, achieves state-of-the-art performance with 70.68\% fewer parameters while exceeding real-time standards, incorporating a new pre-training strategy for sky/cloud segmentation when ImageNet pre-training is unavailable.
    \item A novel SCAM decoder with segregated branches for information processing, enabling precise segmentation while maintaining real-time efficiency.
\end{itemize}

\begin{figure*}[htbp]
    \centering
    \begin{minipage}{0.37\textwidth}
        \centering
        \includegraphics[width=\textwidth]{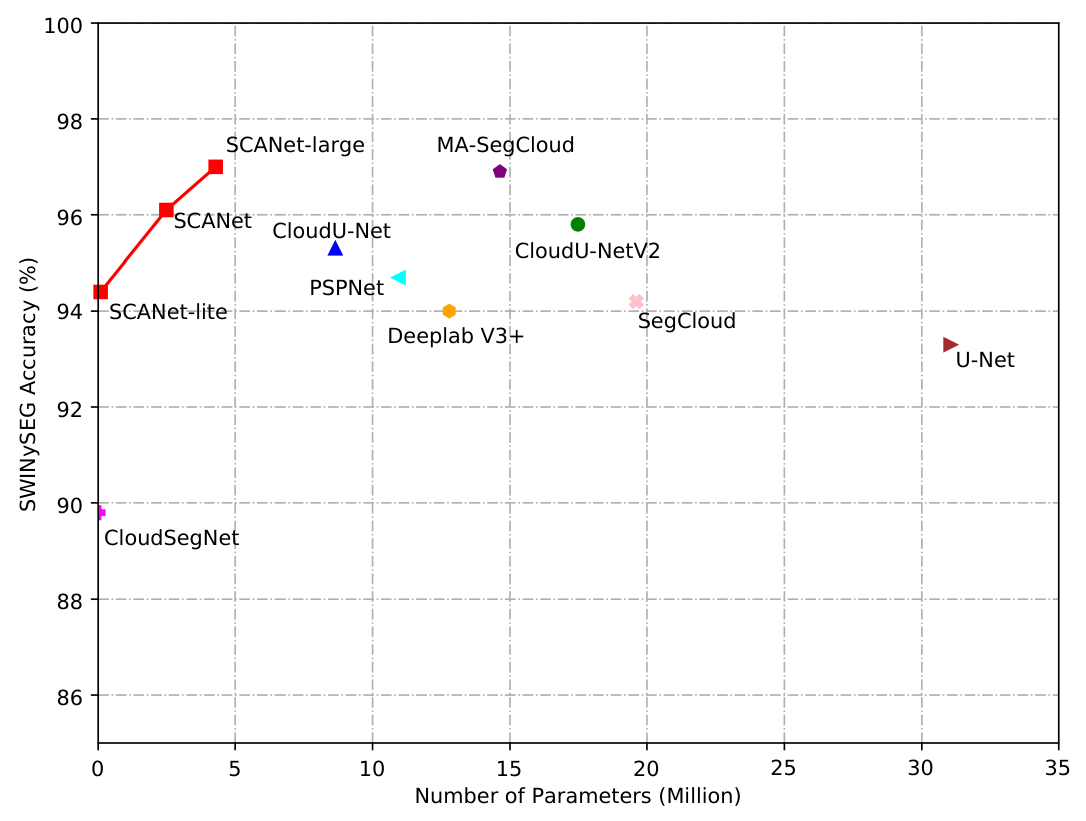}
        \caption{\#Params vs. SWINySEG Accuracy. Our proposed SCANet model successfully achieves a balance between the model size and accuracy. SCANet-large can achieve $97.0\%$ of accuracy in SWINySEG with 4.29 million parameters, while SCANet-lite can achieve $94.4\%$ of accuracy with only 90k parameters.}
        \label{fig:acc_param}
    \end{minipage}
    \hfill
    \begin{minipage}{0.58\textwidth}
        \centering
        \includegraphics[width=\textwidth]{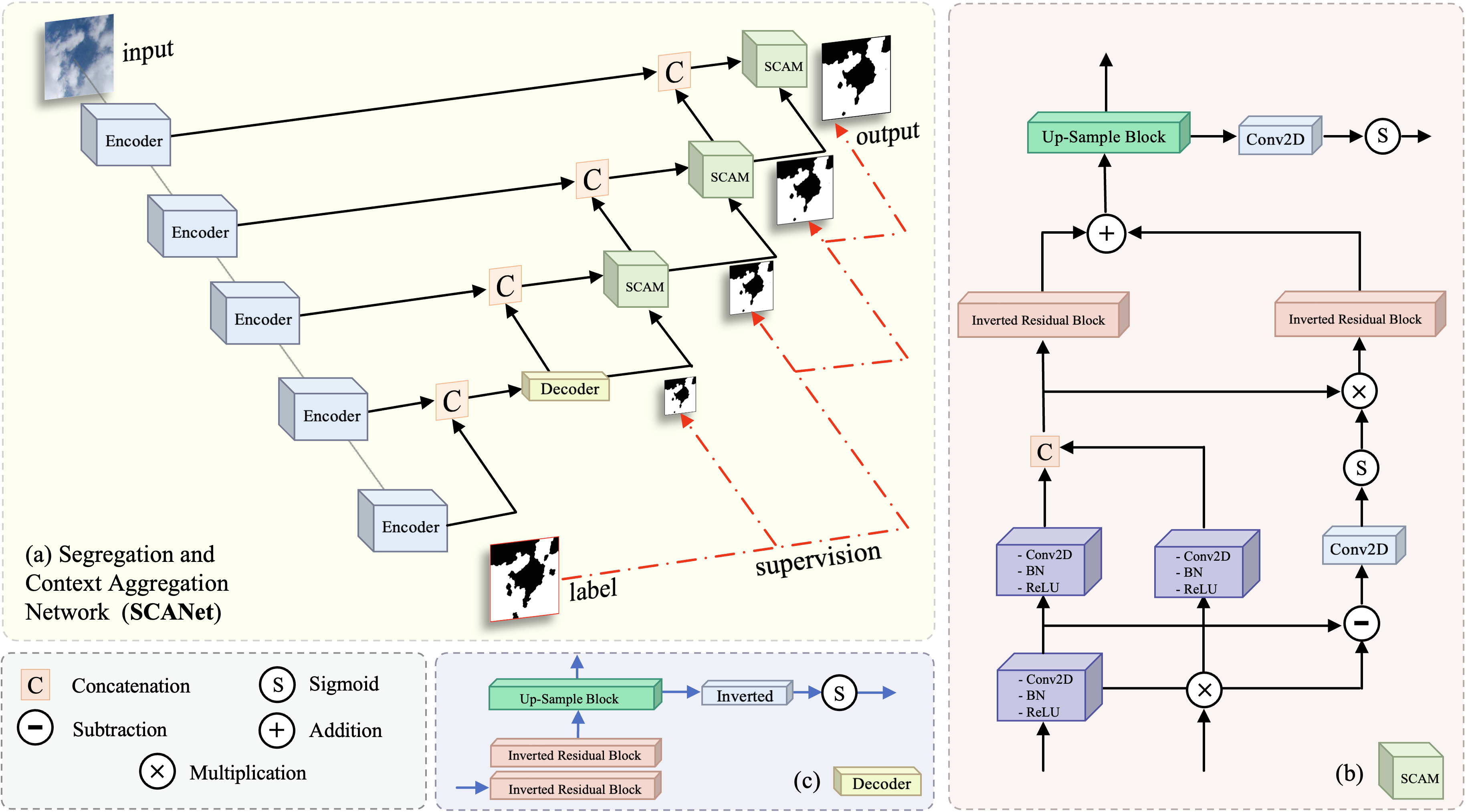}
        \caption{The overall architecture of SCANet and SCAM. (a) presents the pipeline of our SCANet; (b) details the design of our proposed SCAM module; (c) depicts the decoder structure preceding the SCAM modules. Additionally, The architecture of inverted residual block~\citep{sandler2018mobilenetv2} is demonstrated in Fig.~\ref{fig:blocks_compare} in Appendix~\ref{sec:scanet_basic}. The up-sample block consists of an inverted residual block paired with a bilinear up-sample layer.}
        \vspace{-4pt}
        \label{fig:model-structure}
    \end{minipage}
\end{figure*}

\section{SCANet}
\label{sec:format}
Our SCANet is designed based on the U-Net~\citep{ronneberger2015u} structure. We use different backbone networks to extract high-dimensional features in our experiments. The architecture of SCANet is shown in Fig.~\ref{fig:model-structure} (a). We equip our SCANet with MobileNetV2~\citep{sandler2018mobilenetv2} and EfficientNet-B0~\citep{tan2019efficientnet} for different settings. We also propose SWINySEG pre-training (SWPT) for SCANet-lite, demonstrated in Appendix~\ref{sec:appendix_pretrain}.

\subsection{Segregation and Context Aggregation Module (SCAM)}

The Segregation and Context Aggregation Module (SCAM) is a lightweight decoder for sky/cloud segmentation, as shown in Fig.~\ref{fig:model-structure} (b), operating in the 2nd, 3rd, and 4th stages. Given the binary nature of sky/cloud segmentation, SCAM processes these categories separately by first segregating features based on rough segmentation results from the previous stage and then aggregating them. It takes two inputs: a concatenation of the U-Net shortcut and feature maps from the previous layer, $c_{i-1}$, and the segmentation prediction from the prior stage, $s_{i-1}$. The main branch is formulated as $f_{i} = \mathrm{Cat}(\mathrm{Conv}(c_{i-1}), \mathrm{Conv}(c_{i-1} \times s_{i-1}))$, where $f_{i}$ represents foreground (sky) features, and $\mathrm{Cat}$ denotes channel-wise concatenation. The background mask is computed as $m_{i} = \mathrm{Sigmoid}(\mathrm{Conv}(c_{i-1} \times (1 - s_{i-1})))$. To extract background features, we apply the background mask to the core branch, formulated as $b_{i} = f_{i} \times m_{i}$, improving background representation. Finally, element-wise addition aggregates $f_{i}$ and $b_{i}$, producing the output $o_{i}$ as $o_{i} = \mathrm{Conv}(\mathrm{UpSample}(\mathrm{InvRes}(b_{i}) + \mathrm{InvRes}(f_{i})))$, while the stage prediction is obtained through a convolutional layer followed by a sigmoid activation: $s_{i} = \mathrm{Sigmoid}(\mathrm{Conv}({o}_{i}))$.

\subsection{Loss Functions}

In our research work, we use binary cross entropy (BCE) and Intersection of Union (IOU) loss as the loss function in the training of SCANet. These loss functions can be defined as follows:

\vspace{-10pt}
\begin{equation}
	\mathcal{L}_{\mathrm{bce}}(p, y) = - \frac{1}{N} * \sum_{j=1}^{N} (y_{j} * \log{p_{j}} + (1 - y_{j}) * \log{(1-p_{j})})
\end{equation} 
\vspace{-10pt}

\vspace{-10pt}
\begin{equation}
	\mathcal{L}_{\mathrm{iou}}(p, y) = 1 - \frac{1}{n}\sum_{j=1}^{N}(\frac{y_{j}\times p_{j}}{y_{j}+p_{j}-y_{j} \times p_{j}})
\end{equation}
\vspace{-10pt}
then our total training loss under deep supervision can be formulated as:

\vspace{0pt}
\begin{equation}
	\mathcal{L}(p, y) = \sum_{i=1}^{4}\alpha_{i}*(\mathcal{L}_{\mathrm{bce}}(p_{i}, y_{i})+\mathcal{L}_{\mathrm{iou}}(p_{i}, y_{i}))
\end{equation} 
\vspace{- 10pt}

in which $\alpha_{i}$ represents the coefficient of $i$ th SCAM or decoder.

\begin{figure*}[htbp]
	\centering
	\includegraphics[width=0.95\textwidth]{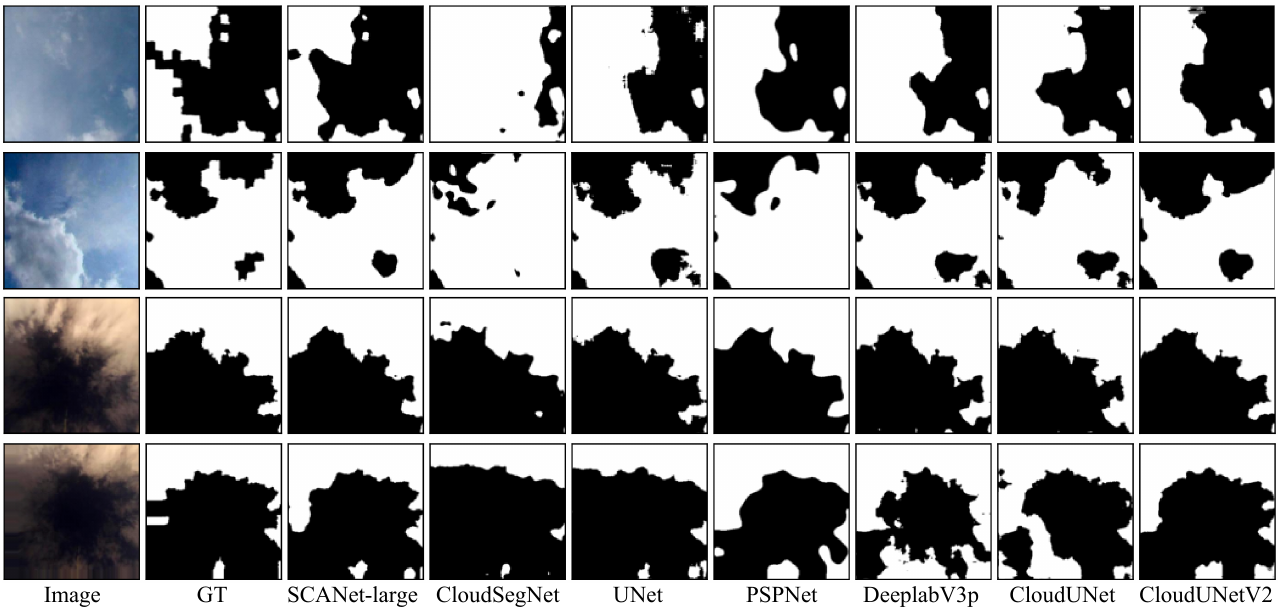}
	\caption{Qualitative comparison of SCANet-large with state-of-the-art approaches on day-time (rows 1–2) and night-time (rows 3–4) images from the SWINySEG dataset.}
	\vspace{-0.3cm}
	\label{fig:res_compare}
\end{figure*}

\section{Datasets \& Configurations}
\subsection{Dataset}
We use the Singapore Whole Sky Nychthemeron Image SEGmentation Database (SWINySEG) as our training dataset, which consists of 6078 day-time and 690 night-time cloud images captured in Singapore using a calibrated camera. Following~\citep{zhang2022novel}, we split the dataset into training and testing sets with a 9:1 ratio. For evaluation, SCANet is tested on three subsets: day-time images (augmented SWIMSEG), night-time images (augmented SWINSEG), and the full SWINySEG dataset. Notably, SCANet is trained only once on the complete SWINySEG dataset.

\subsection{Implementation Details}
We implement SCANet using PaddlePaddle and conduct training on a single NVIDIA Tesla V100-SXM2 16GB GPU. The model is trained for 100 epochs with a batch size of 16. We use the Adam optimizer with an initial learning rate of $1e-3$, $\beta_1 = 0.9$, $\beta_2 = 0.999$, and epsilon set to $1e-8$. The learning rate follows an exponential decay with a decay factor of $\gamma = 0.95$ after each epoch. Evaluation on the test set is conducted every 5 epochs to monitor performance.

As for data augmentation, we apply only random horizontal and vertical flips after resizing the images to a resolution of $320 \times 320$. The augmented images are then scaled to the range $[0, 1]$ and normalized to have a mean of 0.5 across all three channels.

\vspace{-1pt}
\section{Experiments \& Results}
\vspace{-6pt}

\vspace{-1pt}
\subsection{Metrics}
\vspace{-6pt}
In our experiments, we evaluate SCANet using six widely used metrics: accuracy, precision, recall, F-score, error rate, and MIoU. The F-score, which reflects overall model performance, is the harmonic mean of precision and recall, given by \( \frac{2 \times \mathrm{Precision} \times \mathrm{Recall}}{\mathrm{Precision} + \mathrm{Recall}} \). Precision is defined as \( \frac{\mathrm{TP}}{\mathrm{TP}+\mathrm{FP}} \), recall as \( \frac{\mathrm{TP}}{\mathrm{TP}+\mathrm{FN}} \), and error rate as \( \frac{\mathrm{FP}+\mathrm{FN}}{\mathrm{P}+\mathrm{N}} \). Besides, MIoU, a common metric is calculated as \( \mathrm{MIoU} = \frac{\mathrm{miou}_{+} + \mathrm{miou}_{-}}{2} \), where \( \mathrm{miou}_{+} = \frac{\mathrm{TP}}{\mathrm{FN}+\mathrm{FP}+\mathrm{TP}} \) and \( \mathrm{miou}_{-} = \frac{\mathrm{TN}}{\mathrm{TN}+\mathrm{FN}+\mathrm{FP}} \), respectively.

\vspace{-1pt}
\subsection{Qualitative Evaluation}
\vspace{-6pt}
The qualitative comparison of SCANet-large with six prior methods~\citep{dev2019cloudsegnet, shi2020cloudu, shi2021cloudu, ronneberger2015u, Zhao_2017_CVPR, chen2017rethinking} is shown in Fig.~\ref{fig:res_compare}. The leftmost columns present source images and ground truths, with day-time samples in the first two rows and night-time images in the last two. Cloud-sky boundaries in the first row challenge prior methods—CloudSegNet misclassifies sky as cloud, while others miss small patches. SCANet-large, however, accurately segments both. In the second row, it correctly classifies three small sky patches, unlike previous models that confuse sky and cloud. Night-time segmentation is even harder due to limited training data. In the first night-time row, complex cloud structures degrade prior methods' performance, but SCANet-large remains accurate, highlighting its advantages in atmospheric science applications. Additional comparisons are provided in Fig.~\ref{fig:qua_appendix} in Appendix~\ref{sec:appendxeval}.

\vspace{-1pt}
\subsection{Quantitative Evaluation}
\begin{table*}[htbp]
	\centering
        \renewcommand{\arraystretch}{1.3} 
	\caption{Comparison with other state-of-the-art methods on day time, night time, and day+night time images. We highlight the optimal and suboptimal methods in \textbf{bold} and \underline{underline}, respectively. "/" indicates that the corresponding metric was not reported in the original paper.}
    \label{table-1}
	\resizebox{\textwidth}{!}{
	\begin{tabular}{|c|c|c c c c c|c c c c c|c c c c c|}
	    \hline
		\multirow{2}{*}{\textbf{Methods}} & \multirow{2}{*}{\textbf{\#Params}} & \multicolumn{5}{c|}{\textbf{Day-time}}  & \multicolumn{5}{c|}{\textbf{Night-time}} & \multicolumn{5}{c|}{\textbf{Day+Night time}}\\
		\cline{3-17}
		&  & {Acc.} & {Prec.} & {Rec.} & {F1} & {MIoU} & {Acc.} & {Prec.} & {Rec.} & {F1} & {MIoU} & {Acc.} & {Prec.} & {Rec.} & {F1} & {MIoU}\\  
		\hline
		\rowcolor{RowColor} \multicolumn{17}{|c|}{General Semantic Segmentation Models} \\
		\hline
		{U-Net}~\citep{ronneberger2015u} & 31.05M & 0.933 & 0.938 & 0.919 & 0.928 & 0.844 & 0.933 & 0.938 & 0.919 & 0.928 & 0.844 & 0.933 & 0.938 & 0.919 & 0.928 & 0.844\\
		{PSPNet}~\citep{Zhao_2017_CVPR} & 20.95M & 0.947 & 0.949 & 0.935 & 0.942 & 0.873 & 0.947 & 0.950 & 0.935 & 0.942 & 0.873 & 0.947 & 0.950 & 0.935 & 0.942 & 0.873\\
		{DeeplabV3+}~\citep{chen2017rethinking} & 12.80M & 0.940 & 0.941 & 0.920 & 0.936 & 0.860 & 0.940 & 0.941 & 0.932 & 0.936 & 0.860 & 0.940 & 0.941 & 0.932 & 0.936 & 0.860\\
		\hline
		  \rowcolor{RowColor} \multicolumn{17}{|c|}{Special Designed Sky/Cloud Segmentation Models} \\
		\hline
		{CloudSegNet}~\citep{dev2019cloudsegnet} & 0.005M & 0.898 & 0.920 & 0.876 & 0.898 & 0.777 & 0.898 & 0.920 & 0.876 & 0.898 & 0.777 & 0.898 & 0.920 & 0.876 & 0.898 & 0.777\\
		{SegCloud}~\citep{xie2020segcloud} & 19.61M & 0.941 & 0.953 & 0.934 & 0.943 & 0.889 & 0.955 & 0.936 & 0.960 & 0.948 & 0.912 & 0.942 & 0.952 & 0.936 & 0.944 & 0.891\\
            {UCloudNet}~\citep{li2024ucloudnet} & / & 0.940 & 0.920 & 0.940 & 0.930 & / & 0.960 & 0.950 & 0.950 & 0.950 & / & 0.940 & 0.920 & 0.940 & 0.930 & /\\
            {DDUNet}~\citep{li2025ddunet} & 0.33M & 0.953 & 0.953 & / & / & 0.882 & 0.954 & 0.951 & / & / & 0.900 & 0.953 & 0.952 & / & / & 0.884\\
		{CloudU-Net}~\citep{shi2020cloudu} & 8.64M & 0.953 & 0.949 & 0.955 & 0.948 & 0.885 & 0.953 & 0.949 & 0.947 & 0.948 & 0.885 & 0.953 & 0.949 & 0.947 & 0.948 & 0.885\\
		{CloudU-NetV2}~\citep{shi2021cloudu} & 17.48M & 0.958 & 0.955 & 0.952 & 0.953 & 0.895 & 0.958 & 0.955 & 0.952 & 0.953 & 0.895 & 0.958 & 0.955 & 0.952 & 0.953 & 0.900\\
		{MA-SegCloud}~\citep{zhang2022novel} & 14.63M & \underline{0.969} & \underline{0.971} & \textbf{0.970} & \textbf{0.970} & \textbf{0.940} & \underline{0.969} & \underline{0.960} & \textbf{0.970} & \underline{0.965} & \textbf{0.940} & \underline{0.969} & \underline{0.970} & \textbf{0.970} & \textbf{0.970} & \textbf{0.940}\\
		\hline
		\rowcolor{gray!15} {SCANet-lite} & 0.09M & 0.944 & 0.936 & 0.944 & 0.940 & 0.865 & 0.944 & 0.936 & 0.944 & 0.940 & 0.865 & 0.944 & 0.936 & 0.944 & 0.940 & 0.865 \\
		\rowcolor{gray!30} {SCANet} & 2.49M & 0.961 & 0.955 & 0.958 & 0.957 & 0.901 & 0.961 & 0.955 & 0.958 & 0.957 & 0.902 & 0.961 & 0.955 & 0.958 & 0.957 & 0.902 \\
		\rowcolor{gray!45} {SCANet-large} & 4.29M & \textbf{0.970} & \textbf{0.971} & \underline{0.960} & \underline{0.966} & \underline{0.923} & \textbf{0.970} & \textbf{0.971} & \underline{0.960} & \textbf{0.966} & \underline{0.923} & \textbf{0.970} & \textbf{0.971} & \underline{0.960} & \underline{0.966} & \underline{0.923}\\
		\hline
	\end{tabular}
	}
\end{table*}

\begin{table*}[htbp]
    \centering
    \renewcommand{\arraystretch}{1.3}
    \begin{minipage}{0.35\textwidth} 
        \scalebox{0.6}{
        \begin{tabular}{c|c|cccc}
            \toprule
            \multirow{2}{*}{\textbf{Methods}} & \multirow{2}{*}{\textbf{MFlops}} & \multicolumn{2}{c}{\textbf{FP32}} & \multicolumn{2}{c}{\textbf{FP16}}\\
            & & FPS & Latency & FPS & Latency \\ 
            \midrule
            \rowcolor{gray!15} SCANet-lite & 111.216 & 750 & 1.3 ms & 1390 & 0.7 ms\\
            \rowcolor{gray!30} SCANet & 1451.31 & 465 & 2.1 ms & 1124 & 0.8 ms\\
            \rowcolor{gray!45} SCANet-large & 365.85 & 299 & 3.3 ms & 392 & 2.6 ms\\
            \bottomrule
        \end{tabular}
        }
    \end{minipage}
    \hspace{20pt} 
    \begin{minipage}{0.55\textwidth} 
        \caption{MFLops, inference latency, and FPS of SCANets on an NVIDIA Tesla V100 GPU. Models were deployed using TensorRT with FP32 and FP16 precision. Inference latency is measured as the average processing time per image over 1000 inferences.}
        \label{FPS and Infer Time of SCANets}
        \vspace{-10pt}
    \end{minipage}
\end{table*}

Table~\ref{table-1} presents the quantitative evaluation of SCANet on day-time, night-time, and day+night time SWINySEG datasets, comparing it with state-of-the-art methods. We reference~\citep{zhang2022novel} for prior results and ensure a fair comparison by maintaining the same settings. SCANet-large achieves the highest accuracy ($0.970$) and precision ($0.971$), outperforming 8 prior methods, including MA-SegCloud ($0.969$ accuracy, $0.970$ precision), despite having only $4.29$ million parameters—a $70.68\%$ reduction compared to MA-SegCloud ($14.63$ million). The standard SCANet achieves competitive accuracy ($0.960$) and MIoU ($0.900$), while SCANet-lite, with just $0.09$ million parameters, attains $0.945$ accuracy, surpassing larger models like SegCloud ($19.61$ million parameters, $0.942$ accuracy). These results demonstrate SCANet’s efficiency and performance balance, with SCANet-large achieving state-of-the-art accuracy and precision using much fewer parameters. Precision-Recall (PR) curves in Fig.~\ref{pr_fm_curves} in Appendix~\ref{sec:appendxeval} illustrate all methods' overall performance.

\subsection{Ablation Study}
\vspace{-5pt}
To assess our proposed modules, backbone networks, loss functions, and pre-training strategies, we conduct an ablation study shown in Table~\ref{ablation_result}. The baseline U-Net with inverted residual blocks has 0.32M parameters, achieving $92.7\%$ accuracy and $83.2\%$ MIoU with BCE loss. Replacing its backbone with MobileNetV2-lite and adding SCAM (without the Right Branch) boosts accuracy to $93.6\%$ (No. 2) with only 0.09M parameters. The complete SCAM (No. 3) further improves performance, while SWPT provides minor gains. BCE+IOU loss surpasses IOU-only in accuracy and MIoU. Finally, we evaluate MobileNetV2 and EfficientNet-B0 with BCE+IOU loss. To complement quantitative results, Fig.~\ref{fig:ablationres_compare} in Appendix~\ref{sec:appendxeval} provides visualizations of eight key experiments (No. 1, 2, 3, 4, 5, 6, 8, 10), alongside PR and F-Measure curves in Fig.~\ref{ablation_pr_fm_curves} within Appendix~\ref{sec:appendxeval}.

\begin{table*}[htbp]
\renewcommand{\arraystretch}{1.3}
\arrayrulecolor{black} 
	\centering
        \caption{Ablation study on different module compositions, loss functions, backbone networks, and pre-training strategies. SWPT indicates SWINySEG-based pre-training and INPT is ImageNet-based pre-training. We build a light-weight U-Net with 0.32M parameters as the baseline model.}
	\resizebox{\textwidth}{!}{
    	\begin{tabular}{|cccccccccccccc}
    		\cline{1-14}
    		\multicolumn{1}{c|}{\multirow{2}{*}{\textbf{No.}}} & \multicolumn{1}{c|}{\multirow{2}{*}{\textbf{Backbone}}} & \multicolumn{2}{c|}{\textbf{SCAM Configs}} & \multicolumn{2}{c|}{\textbf{Loss Functions}} & \multicolumn{2}{c|}{\textbf{Pre-training}} & \multicolumn{1}{c|}{\multirow{2}{*}{\textbf{\#Params}}} & \multicolumn{5}{c}{\textbf{SWINySEG}} \\
    		\multicolumn{1}{c|}{} & \multicolumn{1}{c|}{} & L Branch & \multicolumn{1}{c|}{R Branch} & BCE & \multicolumn{1}{c|}{IOU} & SWPT & \multicolumn{1}{c|}{INPT} & \multicolumn{1}{c|}{} & Accuracy & Precision & Recall & F-score & MIoU \\ 
    		\cline{1-14}
    		\multicolumn{1}{c|}{1} & \multicolumn{1}{c|}{baseline} & \xmark & \xmark & \cmark & \xmark & \xmark & \multicolumn{1}{c|}{\xmark} & \multicolumn{1}{c|}{\underline{0.32M}} & 0.927 & 0.934 & 0.910 & 0.922 & 0.832\\
    		\Xhline{0.48pt}
    		\rowcolor{RowColor} \multicolumn{1}{c|}{2} & \multicolumn{1}{c|}{MobileNetV2-lite} &  \cmark &  \xmark  & \cmark &  \xmark  &  \xmark  & \multicolumn{1}{c|}{ \xmark } & \multicolumn{1}{c|}{\textbf{0.09M}} & 0.936 & 0.940 & 0.925 & 0.932 & 0.850\\
    		\multicolumn{1}{c|}{3} & \multicolumn{1}{c|}{MobileNetV2-lite} &  \cmark  & \cmark & \cmark &  \xmark  &  \xmark  & \multicolumn{1}{c|}{ \xmark } & \multicolumn{1}{c|}{\textbf{0.09M}} & 0.943 & 0.944 & 0.932 & 0.938 & 0.863\\
           \rowcolor{RowColor}  \multicolumn{1}{c|}{4} & \multicolumn{1}{c|}{MobileNetV2-lite} &  \cmark & \cmark & \cmark &  \xmark  & \cmark & \multicolumn{1}{c|}{ \xmark } & \multicolumn{1}{c|}{\textbf{0.09M}} & 0.944 & 0.943 & 0.933 & 0.938 & 0.863\\
            \multicolumn{1}{c|}{5} & \multicolumn{1}{c|}{MobileNetV2-lite} &  \cmark & \cmark &  \xmark  & \cmark & \cmark & \multicolumn{1}{c|}{ \xmark } & \multicolumn{1}{c|}{\textbf{0.09M}} & 0.940 & 0.941 & 0.934 & 0.938 & 0.856\\
            \rowcolor{RowColor} \multicolumn{1}{c|}{6} & \multicolumn{1}{c|}{MobileNetV2-lite} &  \cmark & \cmark & \cmark & \cmark & \cmark & \multicolumn{1}{c|}{ \xmark } & \multicolumn{1}{c|}{\textbf{0.09M}} & 0.944 & 0.936 & 0.944 & 0.940 & 0.865\\
    		\cline{1-14}
    		\multicolumn{1}{c|}{7} & \multicolumn{1}{c|}{MobileNetV2} &  \cmark & \cmark & \cmark &  \xmark  &  \xmark  & \multicolumn{1}{c|}{ \cmark } & \multicolumn{1}{c|}{2.49M} & 0.960 & 0.958 & 0.951 & 0.955 & 0.898\\
            \rowcolor{RowColor} \multicolumn{1}{c|}{8} & \multicolumn{1}{c|}{MobileNetV2} &  \cmark &  \cmark & \cmark & \cmark &  \xmark  & \multicolumn{1}{c|}{ \cmark } & \multicolumn{1}{c|}{2.49M} & 0.961 & 0.955 & \underline{0.958} & 0.957 & 0.902\\
    	    \cline{1-14}
    	  \multicolumn{1}{c|}{9} & \multicolumn{1}{c|}{EfficientNet-B0} &  \cmark & \cmark & \cmark &  \xmark  &  \xmark  & \multicolumn{1}{c|}{ \cmark } & \multicolumn{1}{c|}{4.29M} & \underline{0.969} & \textbf{0.972} & 0.956 & \underline{0.964} & \underline{0.919}\\
    	    \rowcolor{RowColor} \multicolumn{1}{c|}{10} & \multicolumn{1}{c|}{EfficientNet-B0} &  \cmark & \cmark & \cmark & \cmark &  \xmark  & \multicolumn{1}{c|}{ \cmark } & \multicolumn{1}{c|}{4.29M} & \textbf{0.970} & \underline{0.971} & \textbf{0.960} & \textbf{0.966} & \textbf{0.923}\\
    	    \cline{1-14}
    	\end{tabular}
	}
	\vspace{0.05in}
	\label{ablation_result}
\end{table*}
\vspace{-10pt}

\section{Conclusion}
\vspace{-5pt}
In this paper, we introduce SCANet, a real-time lightweight cloud segmentation model that reduces parameters by 70.9\% while maintaining state-of-the-art performance. The SCANet-large configuration achieves 392 fps in FP16 after TensorRT deployment, whereas SCANet-lite with only 0.09M reaches 1390 fps. We also propose an efficient pre-training strategy that enhances segmentation accuracy when ImageNet pre-training is unavailable. Extensive evaluations with prior advanced methods, confirm SCANet’s superior accuracy and inference speed, exceeding real-time standards.

\textbf{Broader impacts:} SCANet is capable of contributing to progress in atmospheric monitoring and the prediction of extreme weather events. By enhancing cloud segmentation accuracy in real time, it refines climate models while deepening our understanding of cloud dynamics and albedo effects—key factors for forecasting solar irradiance and global warming trends. Its rapid processing facilitates cost-effective, edge-device deployment for localized weather analysis and disaster preparedness. 

\textbf{Future work:} Looking ahead, we intend to extend and enhance our SCANet by developing cloud segmentation-powered downstream tasks. Specifically, our efforts will focus on image-based weather prediction, cloud depth estimation, and broader meteorological studies. We also aim to provide researchers in related areas with SCANet-based automation tools to boost interdisciplinary advancements. To tackle these objectives in resource-limited environments, we will explore transfer learning, semi-supervised learning, and few-shot learning through carefully designed experiments to guarantee the generalization ability in different tasks.

\bibliography{longforms,references}

\newpage
\appendix
\section{Appendix}
\subsection{Related Work}
\label{sec:appendixrelatedwork}
Earlier methods in cloud segmentation relied heavily on traditional techniques, utilizing color features, pixel information, pre-defined convolution filters, and edge detection operators~\citep{long2006retrieving, dev2014systematic, yang2010automatic, liu2014automatic}. These methods, however, often struggled with capturing fine-grained details, resulting in sub-par segmentation accuracy. Additionally, their inability to model probabilistic relationships limited their effectiveness when handling unseen data. 

Since the advent of deep learning, dense feature extractors have been widely used in various domains and applications such as medical prediction~\cite{tang2024optimized, DEV2022100032, pan2024accurate, chen2025mdteamgpt, daniel2025continualdeepactivelearning}, visual perception and calibration~\cite{WANG2022102243,you2025multi, WANG2021stereo, BATRA2022200039, WANGCAR2022}, as well as entertainment trend prediction \cite{xu2025cohesion, xu2024aligngroup, WANG2024100601, xu2024fourierkan, xu2024mentor}. There has also been a significant surge in both environmental meteorology research and remote sensing analyses \citep{10423050, cui2024superpixel, li2025cp2m, cui2024real, 10281791}. In the cloud segmentation field, \citep{dev2019cloudsegnet} developed CloudSegNet, a fully convolutional network that applies down-sampling and up-sampling processes to extract high-dimensional feature maps, resulting in segmentation masks with better boundary accuracy. \citep{dev2019multi} introduced a multi-label segmentation method, classifying cloud images into thin clouds, thick clouds, and sky categories, using a U-Net architecture to enhance segmentation precision. 

In 2021, \citep{shi2020cloudu} presented CloudU-Net, a U-Net-based model that integrates fully connected conditional random field (CRF) layers for refined post-processing. This model was further improved into CloudU-NetV2~\citep{shi2021cloudu}, which included non-local attention~\citep{wang2018non} to better capture long-range dependencies, thus improving segmentation accuracy. Nonetheless, adding multiple attention modules significantly increased computational costs due to additional matrix multiplications. Subsequently, ~\citep{zhang2022novel} introduced MA-SegCloud, which incorporates the convolutional block attention module (CBAM)~\citep{woo2018cbam}, squeeze-and-excitation module (SEM)~\citep{hu2018squeeze}, and asymmetric convolution mechanisms to boost segmentation performance. \citep{amt-15-797-2022} proposed a self-supervised learning approach for semantic cloud segmentation of all-sky images, leveraging pretext tasks like inpainting and DeepCluster to pretrain models on unlabeled data, achieving improved performance with fewer labeled samples.

Recently, Transformer-based models have been explored to handle long-range dependencies in sky/cloud segmentation. ~\citep{liu2022transcloudseg} proposed TransCloudSeg, a hybrid model that merges CNN-based encoders with Transformer-based feature extractors, utilizing a Heterogeneous Fusion Module (HFM) to combine outputs from CNN and Transformer branches, thereby demonstrating superior segmentation results. Additionally, research by ~\citep{de2023residual} and ~\citep{guo2020cloud} underscores the benefits of integrating channel attention mechanisms in U-shaped networks. Building on this concept, ~\citep{buttar2022semantic} extended it by incorporating a U-Net++ ~\citep{zhou2019unet++} architecture with SEM modules, aiming to boost feature extraction capabilities. ~\citep{partio2024cloudcast} proposed CloudCast, a U-Net-based model for total cloud cover nowcasting. Trained on five years of satellite data, it outperforms numerical weather prediction models and enhances short-term cloud forecasting.

With the growing emphasis on lightweight models, research has increasingly focused on balancing performance and efficiency in detection and segmentation tasks \citep{li2023daanet, Li_2024_BMVC, wang2024airshot}. In cloud segmentation, Li \textit{et al.} proposed UCloudNet~\citep{li2024ucloudnet}, which utilizes residual connections within U-Net to stabilize training in lightweight models. More recently, in 2025, Li \textit{et al.} introduced DDUNet~\citep{li2025ddunet}, incorporating weighted dilated convolution and a dynamic weight and bias generator to further enhance performance in compact architectures. Despite these advances, balancing segmentation accuracy and computational efficiency remains a persistent challenge. Our proposed SCANet aims to tackle this issue by leveraging lightweight architectures while maintaining high segmentation performance.

Cloud segmentation models are also capable of boosting the downstream tasks, such as time series forecasting \citep{qiu2025duet, qiu2025easytime, AutoCTS++, qiu2025comprehensive, li2024foundts, qiu2024tfb, gao2024diffimp, huo2025ct} and anomaly detection \citep{wu2024catch, hu2024multirc, huo2025enhancing}, which opens new direction for improved rain forecasting and severe weather detection.

\subsection{SCANet - Basic Building Blocks}
\label{sec:scanet_basic}
The fundamental building blocks in computer vision deep learning-based tasks can be categorized as derivatives of three key modules: straight-forward structures, residual blocks~\citep{he2016deep}, and inverted residual blocks~\citep{sandler2018mobilenetv2}. Fig.~\ref{fig:blocks_compare} (a) illustrates the straight-forward structure, which consists of a $3\times3$ convolution layer, a batch normalization layer, and a ReLU activation. This structure was widely used in early CNN models. Fig.~\ref{fig:blocks_compare} (b) presents the residual block architecture, which introduces a shortcut connection to facilitate the training of deep CNN networks. In contrast, Fig.~\ref{fig:blocks_compare} (c) depicts the inverted residual block, which first expands channels using a series of Conv2D-BatchNorm-ReLU6 layers. The features then pass through another set of similar layers, except that the standard Conv2D operation is replaced by depth-wise convolution (DWConv in Fig.~\ref{fig:blocks_compare}), where the number of groups is set to the number of input channels, significantly reducing parameter count. Finally, a $1\times1$ convolution layer and a batch normalization layer reduce the channels back to the input dimension before applying element-wise addition.

Notably, the inverted residual block utilizes ReLU6 instead of ReLU, as ReLU6 caps the output value at 6, preventing excessively large activations. This bounded range helps maintain accuracy when performing inference in lower precision settings (e.g., FP16 mode).

\begin{table*}[htbp]
    \centering
    \renewcommand{\arraystretch}{1.3}
    \begin{minipage}{0.35\textwidth} 
        \centering
        \includegraphics[width=\linewidth]{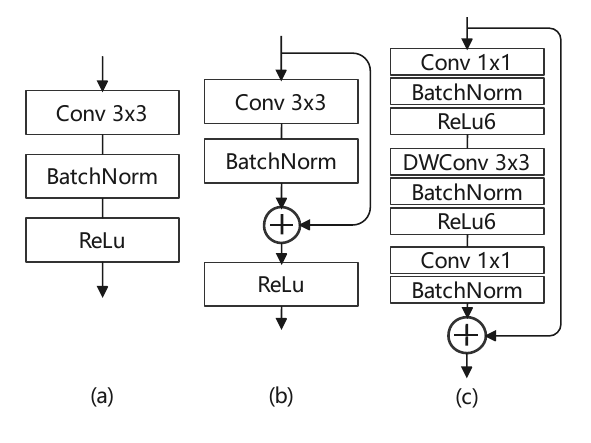} 
    \end{minipage}
    \hspace{-5pt} 
    \begin{minipage}{0.6\textwidth} 
        \caption{Comparison of basic building blocks widely used in backbone networks. (a) shows the straightforward structure with a simple convolutional layer followed by batch normalization and activation, commonly employed in early CNN architectures. (b) illustrates the residual block~\citep{he2016deep} (c) demonstrates the inverted-residual block, designed to reduce parameters and computational cost through channel expansion and depth-wise separable convolutions~\citep{sandler2018mobilenetv2}.}
        \label{fig:blocks_compare}
        \vspace{-10pt}
    \end{minipage}
\end{table*}

\begin{figure*}[htbp]
	\centering
	\includegraphics[width=0.95\textwidth]{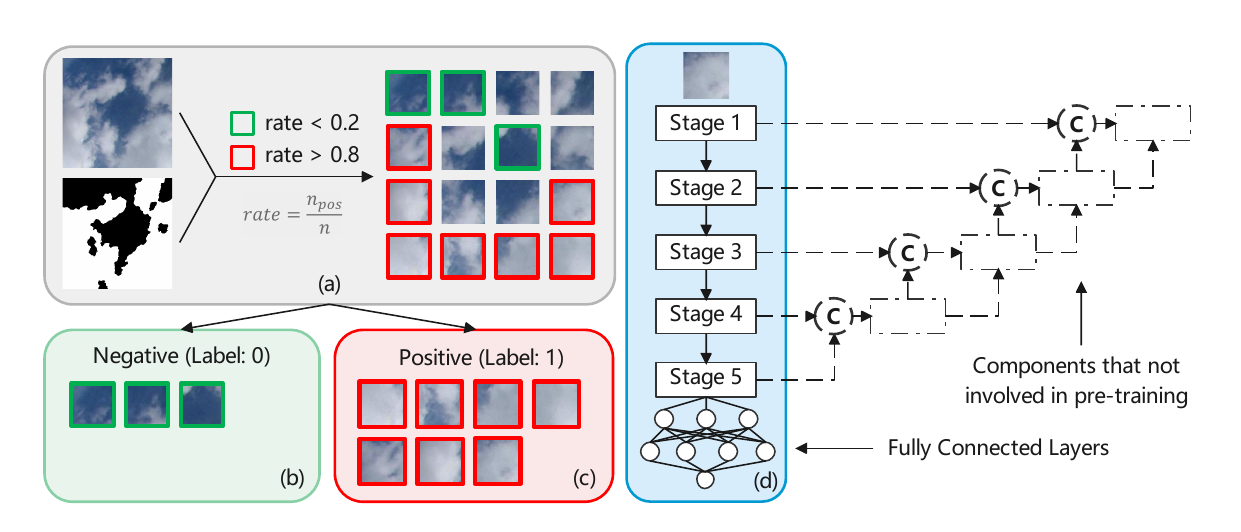}
	\caption{Schematic diagram of SWINySEG-based pre-training. (a) illustrates the positive and negative sample generation process. (b) indicates the negative samples. (c) is positive samples. (d) represents the modules involved in pre-training.}
	\vspace{-0.3cm}
	\label{fig:pre-training}
\end{figure*}

\subsection{SWINySEG-based Pre-training (SWPT)}
\label{sec:appendix_pretrain}
Pre-training is widely used in computer vision tasks to improve performance and accelerate convergence, particularly for complex tasks such as semantic segmentation and object detection. Since training a model from scratch is computationally expensive, pre-training typically involves training the backbone on ImageNet~\citep{deng2009imagenet} before fine-tuning on the target dataset. In SCANet and SCANet-large, we directly reuse pre-trained weights, as no modifications are made to the backbone. However, SCANet-lite introduces architectural changes that require pre-training from scratch. Given the high cost and time required for ImageNet pre-training, we propose an alternative strategy leveraging the SWINySEG dataset, as illustrated in Fig.~\ref{fig:pre-training}. 

Our approach involves iterating through the SWINySEG dataset, splitting each image into $16$ patches, and assigning labels based on the proportion of cloud pixels in each patch:

\begin{equation} 
    \mathrm{rate} = \frac{n_{\mathrm{pos}}}{n} 
\end{equation}
where $n_{\mathrm{pos}}$ represents the number of cloud pixels (label 1), and $n$ denotes the total number of pixels in the patch. If $\mathrm{rate} > 0.8$, the patch is labeled as a positive sample (cloud); if $\mathrm{rate} < 0.2$, it is labeled as a negative sample (sky). Patches with $\mathrm{rate}$ between $0.2$ and $0.8$ are ignored to ensure clear separation between classes. This threshold selection balances the number of positive and negative samples. This threshold selection helps to mitigate ambiguous regions, thus improving the clarity of the pre-training labels. By focusing only on well-defined cloud and sky regions, this strategy enhances the quality of feature representations learned by the model. During pre-training, we remove all decoders and their connections to the backbone, replacing them with a fully connected layer to facilitate feature learning.

\subsection{Additional Experiment}
\label{sec:appendxeval}
\begin{figure}[htbp]
    \centering
	\includegraphics[height=1.7in]{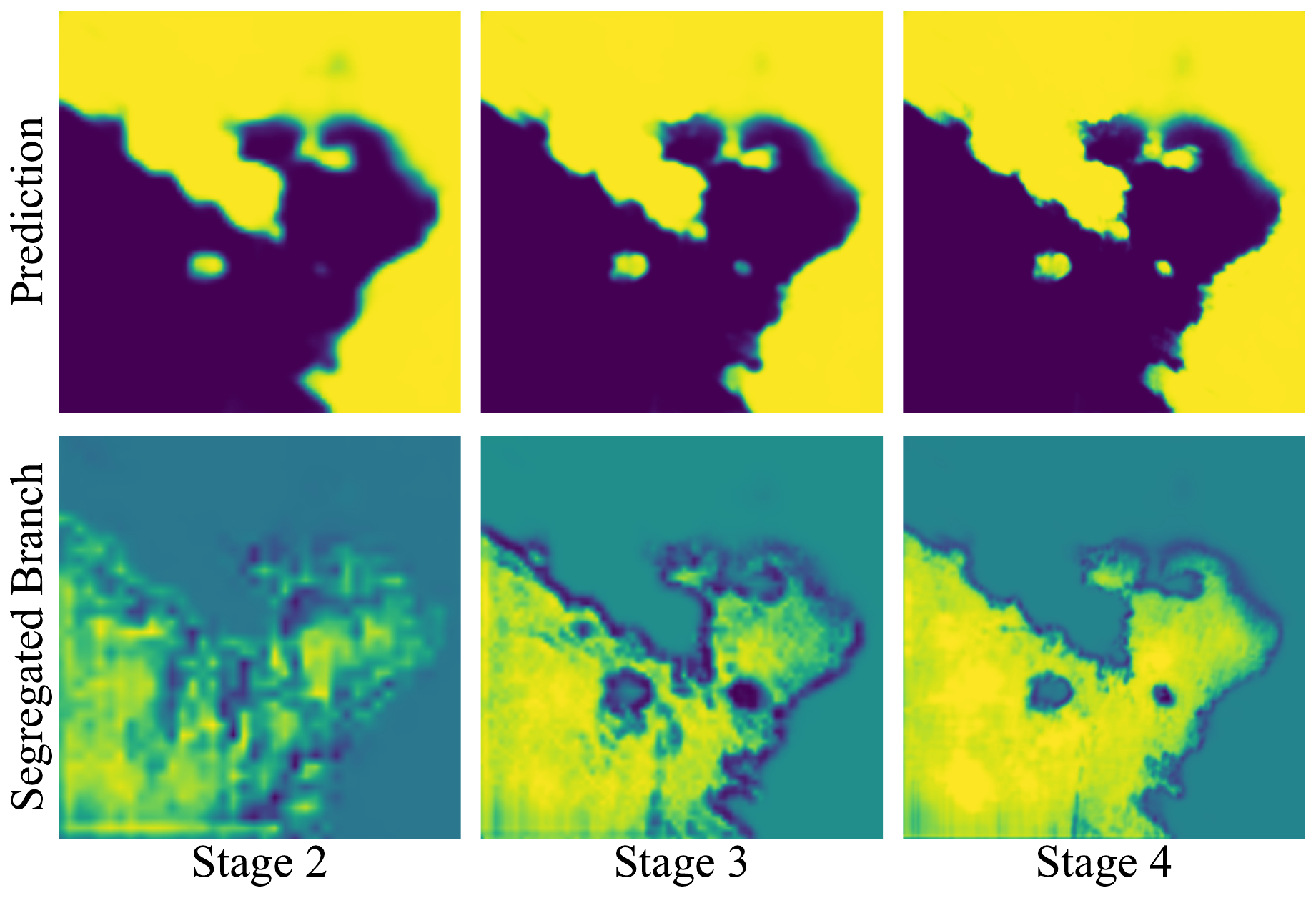}
	\caption{Visualization of SCAM output $s_{i}$ (first row) and background mask $m_{i}$ (second row)}
	\label{fig:SCAM_vis}
\end{figure}

\begin{figure*}[htbp]
    \centering
	\includegraphics[width=0.95\textwidth]{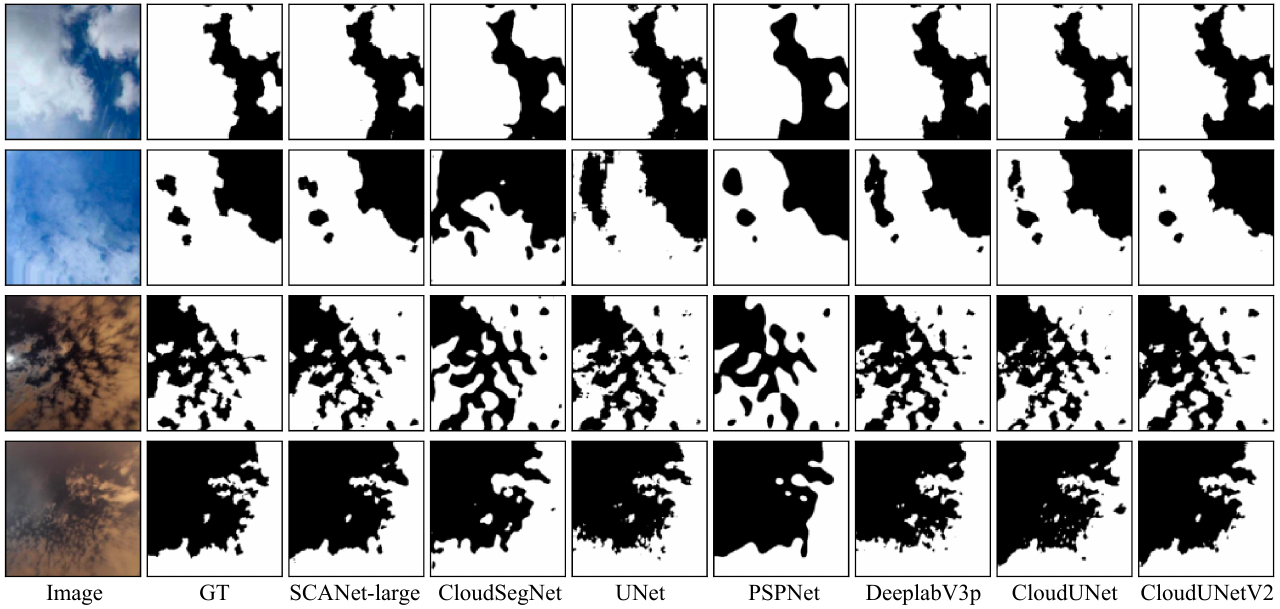}
	\caption{Additional qualitative experiments comparing SCANet-large with state-of-the-art approaches on daytime (rows 1–2) and nighttime (rows 3–4) images from the SWINySEG Dataset}
	\label{fig:qua_appendix}
\end{figure*}

We assess the effectiveness of the SCAM by visualizing its intermediate outputs, as illustrated in Fig.~\ref{fig:SCAM_vis}. The resolutions of the segregated branch output (background mask) from stages 2 to 4 are $40 \times 40$, $80 \times 80$, and $160 \times 160$, while the corresponding stage predictions have resolutions of $80 \times 80$, $160 \times 160$, and $320 \times 320$. At stage 2, the background mask is coarse but already outlines the general segmentation shape. By stage 3, the mask is refined with sharper boundaries, leveraging information from the previous stage. In the final stage, the background mask and stage prediction become well-defined, appearing nearly pure yellow in their combination. This confirms SCAM's effectiveness, particularly in enhancing feature utilization through element-wise operations (addition, multiplication, subtraction) and sigmoid activation—without requiring learnable parameters.

\begin{figure*}[htbp]
    \centering
    \begin{subfigure}[b]{0.32\textwidth}
        \includegraphics[width=\textwidth]{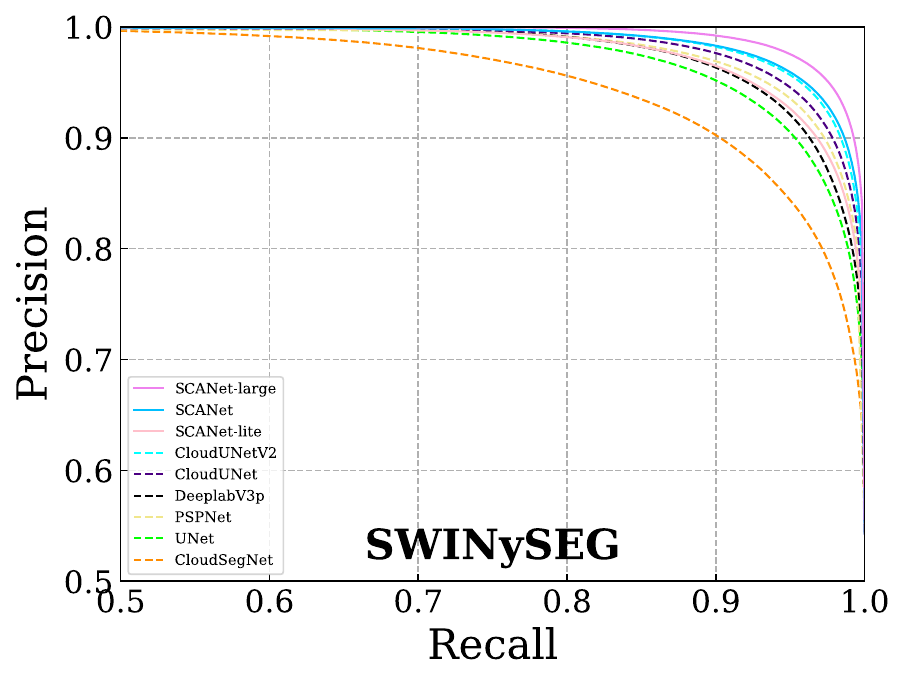}
        \caption{SWINySEG PR curve}
    \end{subfigure}
    \begin{subfigure}[b]{0.32\textwidth}
        \includegraphics[width=\textwidth]{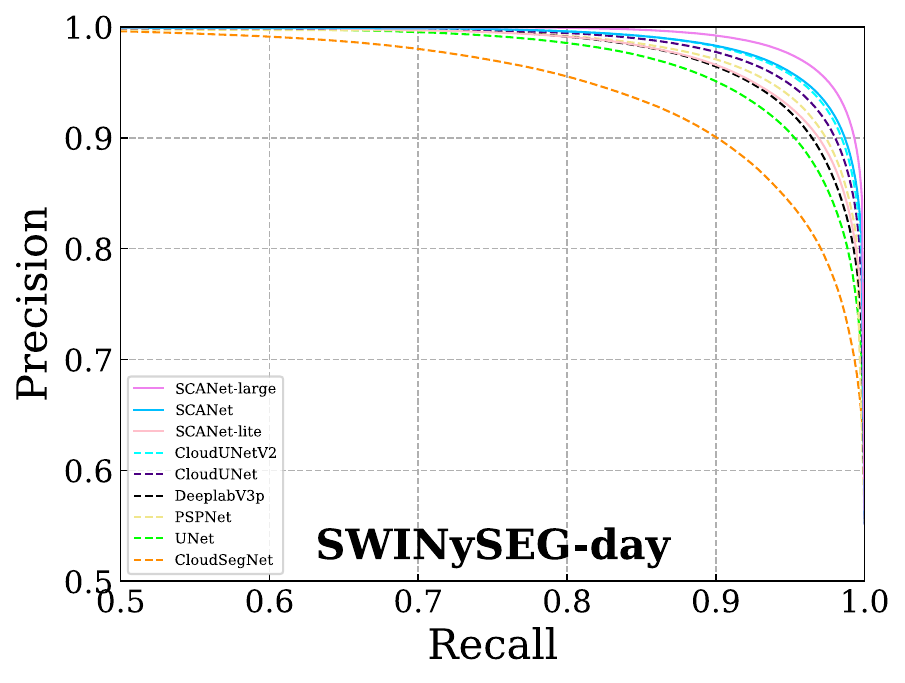}
        \caption{SWINySEG-day PR curve}
    \end{subfigure}
    \begin{subfigure}[b]{0.32\textwidth}
        \includegraphics[width=\textwidth]{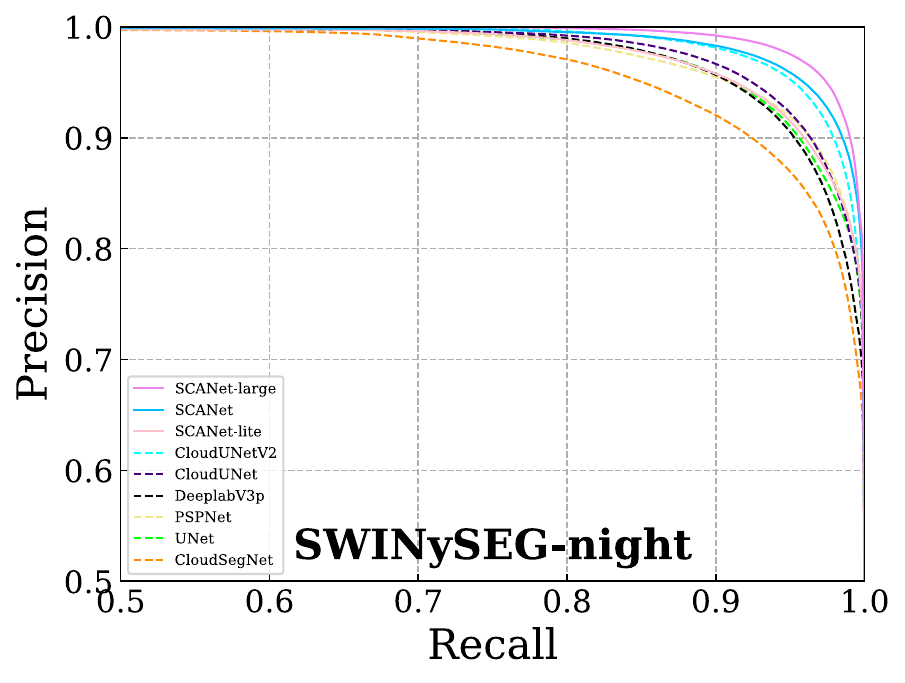}
        \caption{SWINySEG-night PR curve}
    \end{subfigure}
    \begin{subfigure}[b]{0.32\textwidth}
        \includegraphics[width=\textwidth]{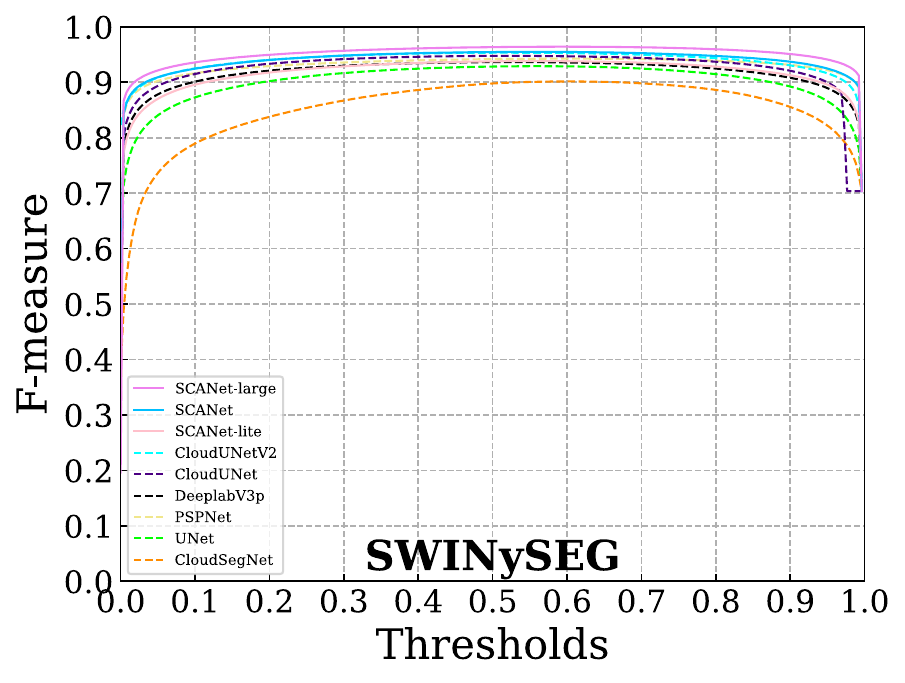}
        \caption{SWINySEG F-measure}
    \end{subfigure}
    \begin{subfigure}[b]{0.32\textwidth}
        \includegraphics[width=\textwidth]{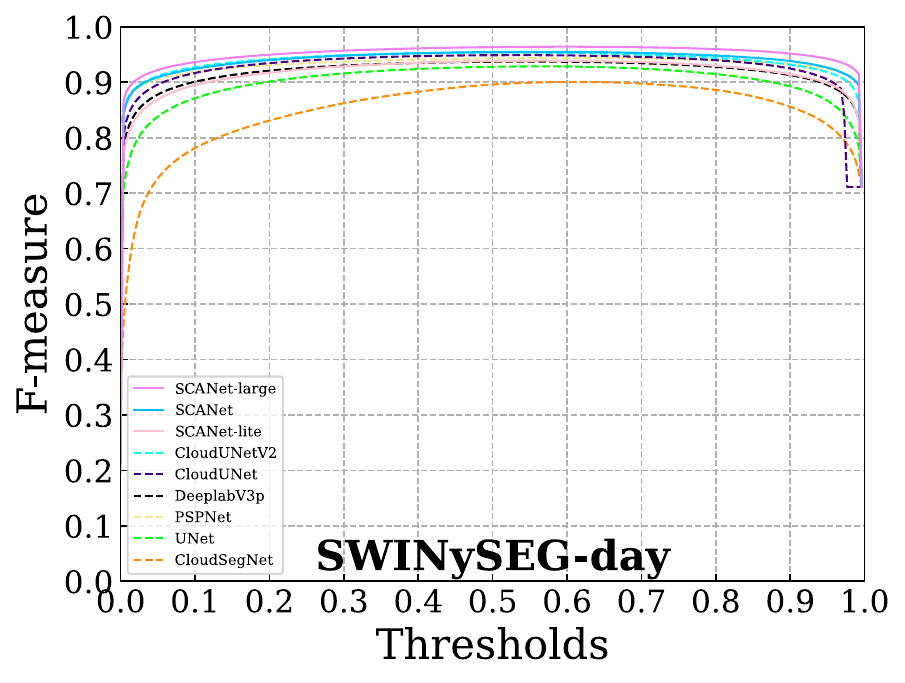}
        \caption{SWINySEG-day F-measure}
    \end{subfigure}
    \begin{subfigure}[b]{0.32\textwidth}
        \includegraphics[width=\textwidth]{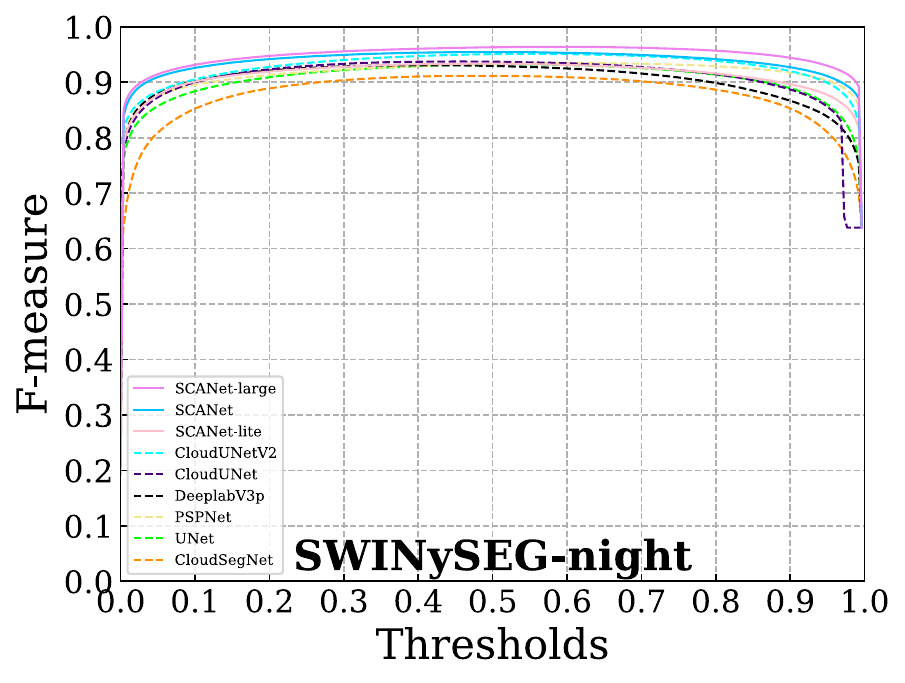}
        \caption{SWINySEG-night F-measure}
    \end{subfigure}
    \caption{PR curves (first row) and F-measure curves (second row) on SWINySEG, SWINySEG-day, and SWINySEG-night dataset.}
    \label{pr_fm_curves}
\end{figure*}

\begin{figure*}[htbp]
    \centering
    \begin{subfigure}[b]{0.32\textwidth}
        \includegraphics[width=\textwidth]{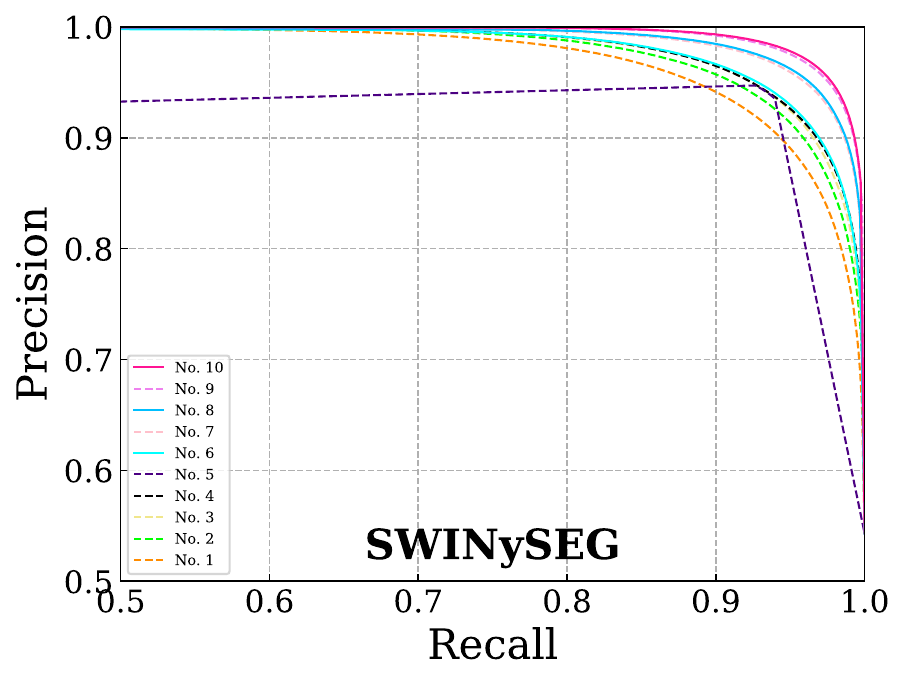}
        \caption{SWINySEG PR curve}
    \end{subfigure}
    \begin{subfigure}[b]{0.32\textwidth}
        \includegraphics[width=\textwidth]{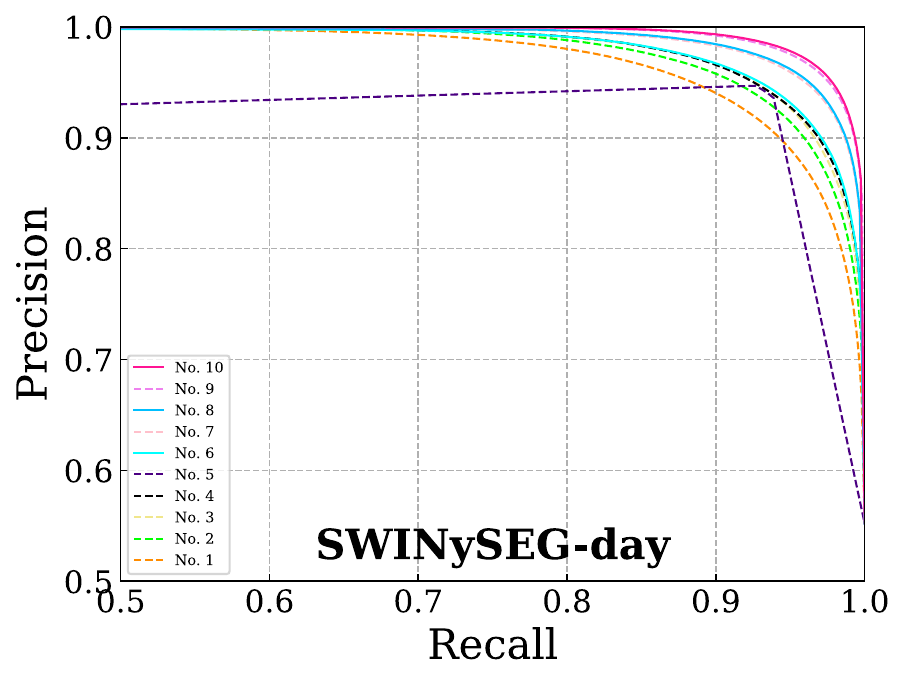}
        \caption{SWINySEG-day PR curve}
    \end{subfigure}
    \begin{subfigure}[b]{0.32\textwidth}
        \includegraphics[width=\textwidth]{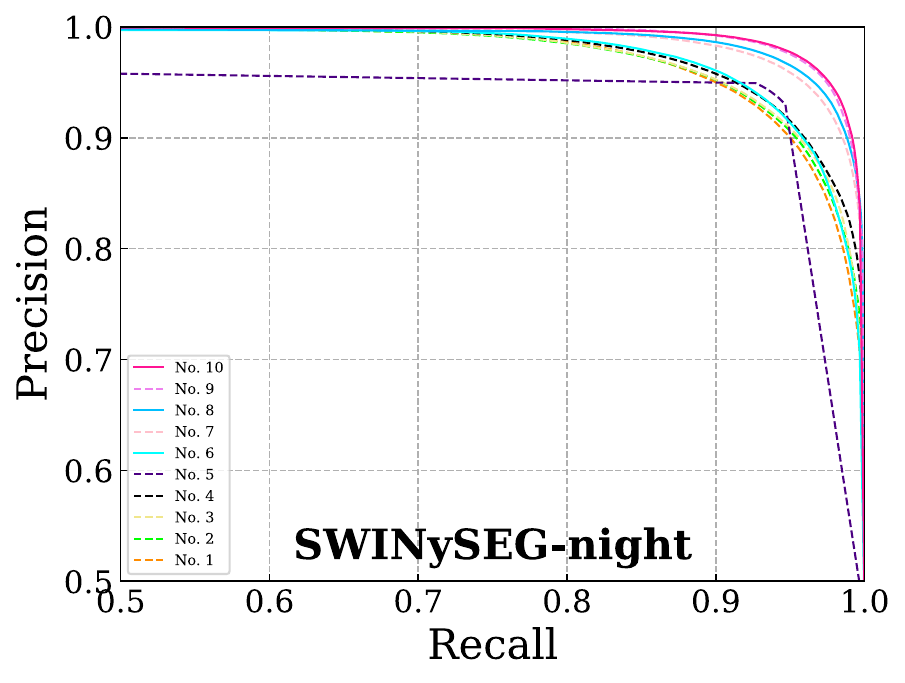}
        \caption{SWINySEG-night PR curve}
    \end{subfigure}
    \begin{subfigure}[b]{0.32\textwidth}
        \includegraphics[width=\textwidth]{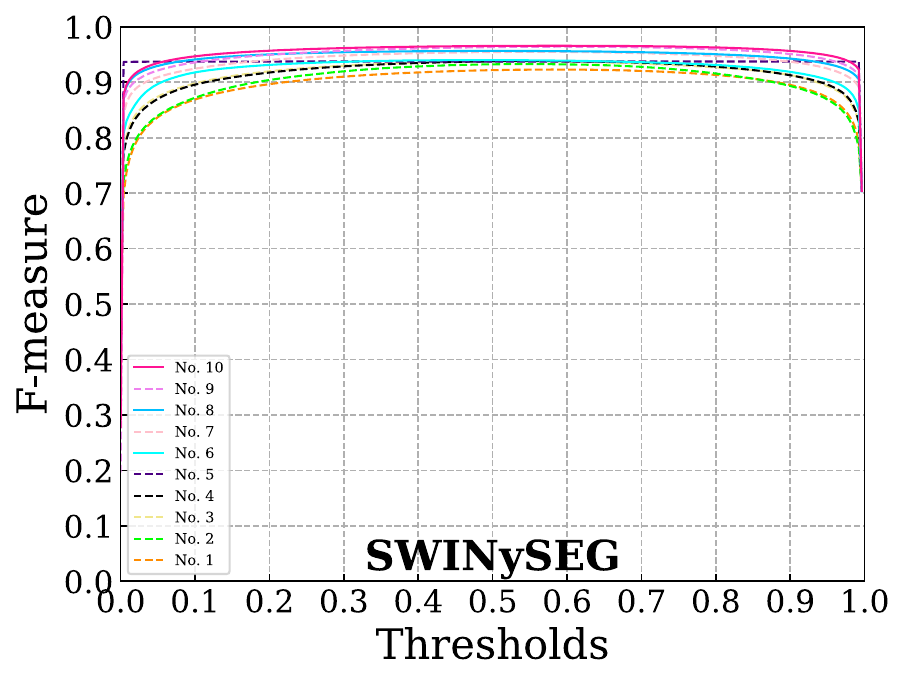}
        \caption{SWINySEG F-measure}
    \end{subfigure}
    \begin{subfigure}[b]{0.32\textwidth}
        \includegraphics[width=\textwidth]{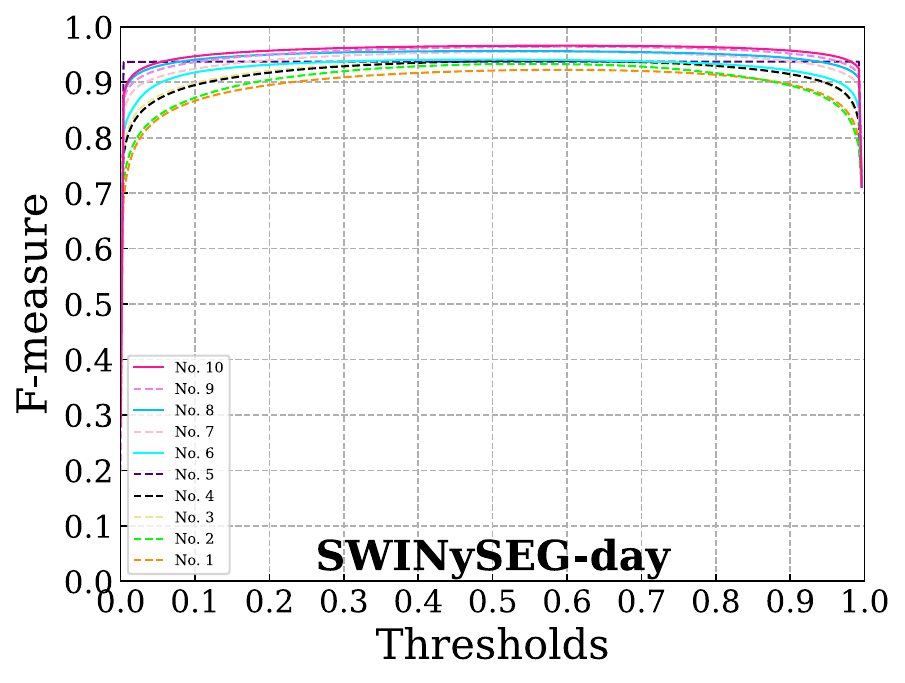}
        \caption{SWINySEG-day F-measure}
    \end{subfigure}
    \begin{subfigure}[b]{0.32\textwidth}
        \includegraphics[width=\textwidth]{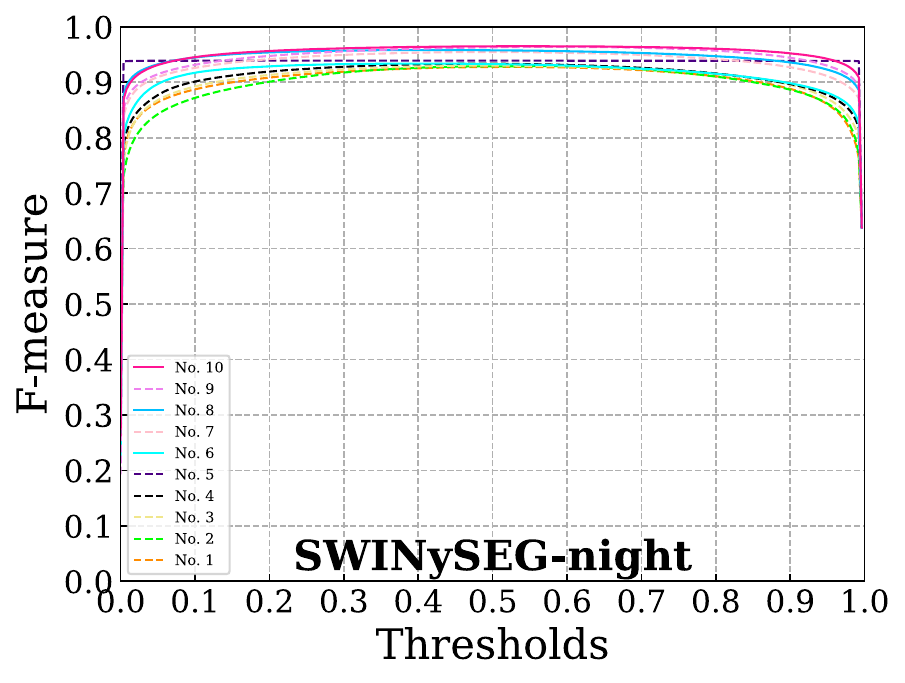}
        \caption{SWINySEG-night F-measure}
    \end{subfigure}
    \caption{PR curves (top row) and F-measure curves (bottom row) for the ablation study results in Table~\ref{ablation_result} on SWINySEG, SWINySEG-day, and SWINySEG-night.}
    \label{ablation_pr_fm_curves}
\end{figure*}

\begin{figure*}[htbp]
	\centering
	\includegraphics[width=0.95\textwidth]{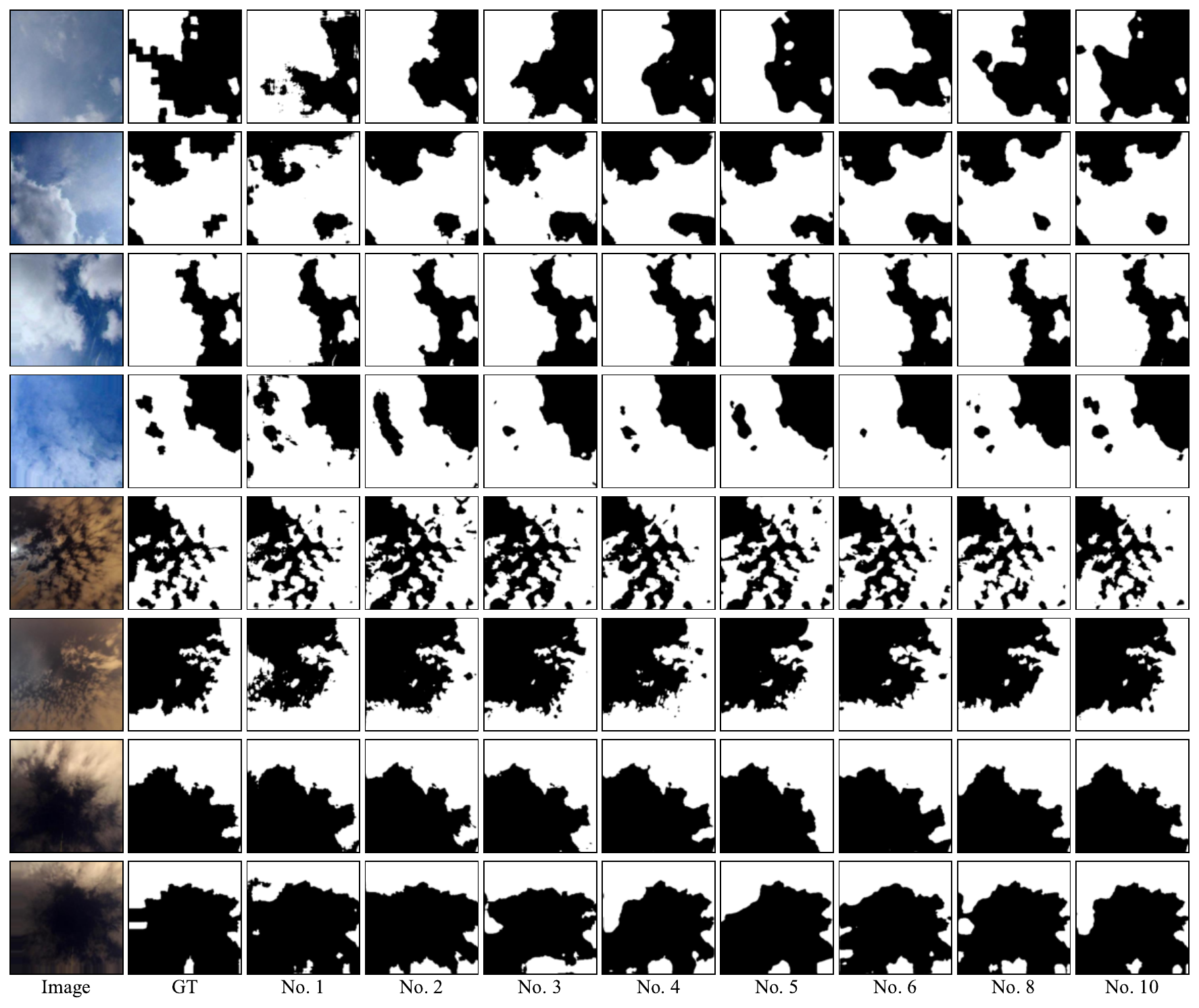}
	\caption{Qualitative visualization of ablation study on day-time and night-time images of SWINySEG dataset; This figure show the prediction maps of experiments No. 1, No. 2, No. 3, No. 4, No. 5, No. 6, No. 8, and No. 10 of Table~\ref{ablation_result}}
	\vspace{-0.3cm}
	\label{fig:ablationres_compare}
\end{figure*}

We provide Fig. \ref{fig:qua_appendix} to complement the qualitative experiment and Fig. \ref{pr_fm_curves} to supplement the quantitative experiment. In addition, to complement the ablation study, we provide Precision-Recall (PR) and F-measure curves for SCANet-large and six prior methods, including CloudSegNet~\citep{dev2019cloudsegnet}, CloudUNet~\citep{shi2020cloudu}, CloudUNetv2~\citep{shi2021cloudu}, U-Net~\citep{ronneberger2015u}, PSPNet~\citep{Zhao_2017_CVPR}, and DeepLabV3plus~\citep{chen2017rethinking}, as shown in Fig.~\ref{ablation_pr_fm_curves}. The PR curve illustrates the balance between precision and recall across various thresholds, while the F-measure curve highlights the model's performance at different thresholds, further validating SCANet's effectiveness in cloud segmentation. We also provided Fig.~\ref{fig:ablationres_compare} showing visualizations of eight essential experiments (No. 1, 2, 3, 4, 5, 6, 8, 10). 

\subsection{Discussion}
Ground-based sky/cloud segmentation extracts cloud structures from Earth-based observations, enabling cloud distribution visualization and supporting downstream meteorological applications, such as weather forecasting and anomaly detection. Deep learning significantly enhances accuracy and efficiency in this task. Early methods, like CloudSegNet~\citep{dev2019cloudsegnet}, employ simple encoder-decoder architectures with convolution and max-pooling layers, ensuring computational efficiency but often falling short in accuracy for advanced meteorological analysis. To enhance performance, many approaches incorporate larger backbone networks or non-local attention mechanisms~\citep{wang2018non}, which improve feature extraction but substantially increase computational complexity.

SCANet introduces a new strategy to optimize both accuracy and efficiency by utilizing lightweight backbone networks while refining decoder design through SCAM. SCAM employs two branches—Left and Right—to process features with high sky and cloud weights, respectively, leveraging prior decoder predictions. Their outputs are then combined to generate the final segmentation. As demonstrated in ablation experiments (No. 1, No. 2, No. 3) in Table~\ref{ablation_result}, both branches contribute significantly to segmentation performance. This design ensures balanced consideration of sky and cloud regions. Additionally, depth-wise convolution is used to minimize decoder parameters, effectively maintaining high segmentation accuracy while reducing computational cost.

\end{document}